\documentclass[conference,10]{IEEEtran}
\IEEEoverridecommandlockouts
\usepackage{minibox}
\IEEEpubid{\makebox[\columnwidth]{\minibox{Submitted to IEEE/ACM Third Workshop on Deep \\  Learning on Supercomputers (DLS)~\copyright~2019 IEEE }} \hspace{\columnsep}\makebox[\columnwidth]{ }}
\usepackage[english]{babel}
\usepackage[utf8x]{inputenc}
\usepackage[T1]{fontenc}

\usepackage[a4paper,top=3cm,bottom=2cm,left=3cm,right=3cm,marginparwidth=1.75cm]{geometry}

\usepackage{mathtools}
\usepackage{amssymb}
\usepackage{graphicx}
\usepackage{color}
\usepackage{subcaption}
\usepackage{float}
\usepackage[ruled,noline]{algorithm2e}
\usepackage[colorinlistoftodos]{todonotes}
\usepackage[colorlinks=true, allcolors=blue]{hyperref}

\title{DC-S3GD: \\ Delay-Compensated Stale-Synchronous SGD for Large-Scale Decentralized Neural Network Training}

\author{\IEEEauthorblockN{Alessandro Rigazzi}
        \IEEEauthorblockA{Cray Switzerland \\Hochbergerstr. 60C\\ 4057 Basel, Switzerland \\Email: arigazzi@cray.com}}

\begin{document}
\maketitle
\IEEEpubidadjcol

\begin{abstract}
Data parallelism has become the \textit{de facto} standard for training Deep Neural Network on multiple processing units.
In this work we propose DC-S3GD, a decentralized (without Parameter Server) stale-synchronous version of the Delay-Compensated Asynchronous Stochastic Gradient Descent (DC-ASGD) algorithm. In our approach, we allow for the overlap of computation and communication, and compensate the inherent error with a first-order correction of the gradients. We prove the effectiveness of our approach by training Convolutional Neural Network with large batches and achieving state-of-the-art results.

\end{abstract}
\vspace{2.0\baselineskip}

\section{Introduction}
\label{sec:introduction}
Training Deep Neural Networks (DNNs) is a time- and resource-consuming problem. For example, to train a DNN to state-of-the-art accuracy on a single processing unit, the total time needed is in the order of magnitude of days, or even weeks \cite{MLPerf}. For this reason, in recent years, several algorithms have been developed to allow users to perform parallel or distributed training of DNNs \cite{Gupta7837841}. With the correct use of parallelism, training times can be reduced down to hours, or even minutes, \cite{MLPerf,goyal2017accurate,krizhevsky2014weird,you2017large}. The reader interested in a broad survey of Deep Learning algorithms is referred to \cite{DBLP:journals/corr/abs-1802-09941}, which is also a great resource for taxonomy and classification of different parallel training strategies.

The most widely adopted type of training parallelism, and the one we will employ in this work, is denominated \textit{data parallelism}: the DNN is replicated on different Processing Units, each replica is trained on a subset of the training data set, and updates (usually in the form of gradients) are regularly aggregated, to create a single update which is then applied to all the DNN replicas.
The way updates are aggregated differs across algorithms in terms of communication scheme, distribution of roles among processing units, and message frequency and content. We will discuss different approaches and architectures in Section~\ref{sec:related_work}.

In Section~\ref{sec:dcdssgd} we describe our approach, which constitutes a modification to the DC-ASGD algorithm proposed in \cite{DBLP:journals/corr/ZhengMWCYML16}. Our approach shows promising results for Convolutional Neural Networks (CNNs): in Section~\ref{sec:results}  we report the results obtained when training different networks on the well-known ImageNet-1k data set, which has imposed itself as the standard benchmark for CNN performance assessment.

In Section~\ref{sec:future_work} we propose possible extensions to the presented algorithm, and outline what advantages they could bring.

\section{Related Work}
\label{sec:related_work}
With the growing availability of parallel systems, such as clusters and supercomputers, both as on-premises or cloud solutions, the demand for fast, reliable, and efficient parallel training scheme has been fueling research in the Artificial Intelligence community \cite{DBLP:journals/corr/abs-1802-09941,goyal2017accurate,krizhevsky2014weird,you2017large,ma2017accelerated}. The most widespread technique, data parallelism, can be applied to many different areas, such as image classification, Reinforcement Learning, or Natural Language Processing \cite{openai2018empirical}. 
When data-parallel training has to be scaled to large systems, convergence problems and loss of generalization arise from the fact that the global batch size becomes very large \cite{Li:2018:VLL:3327345.3327535,smith2017bayesian,openai2018empirical}.

As suggested in \cite{DBLP:journals/corr/abs-1802-09941}, data-parallel training methods can be classified according to two independent aspects: synchronicity (or model consistency across different processes) and communication topology (centralized or decentralized). Synchronous methods are those which ensure that after each training iteration each process (or worker) holds a copy of exactly the same weights; asynchronous methods allow workers to get out of date, receiving updated weights only when they request them (usually after having computed a local update). Centralized communication schemes imply the existence of so-called Parameter Servers, processes which have the task of collecting  weight gradients from workers, and send back updated weights; in decentralized schemes, each worker participates in collective communications to compute the weight updates, e.g. via MPI all-reduce calls.  

\subsection{Advantages and Disadvantages of Different Training Schemes}
\label{sec:advantages}
Historically, when the first major Deep Learning toolkits (such as e.g., TensorFlow \cite{tensorflow2015-whitepaper} or MXNet \cite{chen2015mxnet}) started offering the possibility of parallel training, they did so by implementing techniques with centralized communication, i.e. with Parameter Servers (PSs). As every centralized communication scheme, the PS-paradigm does not scale efficiently. With a growing number of Workers, PSs become bottlenecks, and communication becomes of the many-to-few type. Nevertheless, asynchronous methods often use this paradigm, as it allows workers to send updates independently, without waiting for other workers to complete processing their batches. The most straight-forward algorithm for this setting is clearly the Asynchronous SGD, which has been improved during years with respect to many aspects \cite{DBLP:journals/corr/WangGCLY17,keuper2015asynchronous,DBLP:journals/corr/ZhengMWCYML16}, but its core mechanism can be summarized as follows:
\begin{itemize}
    \item at the beginning of the computation, every worker receives an exact copy of the weights from the PSs
    \item every worker processes a mini-batch and sends the computed gradients to the PSs, which apply them to their local copy of the weights, and send the updated weights to the worker which initiated the communication
    \item the worker proceeds to process another batch, while the PSs wait for gradients from other workers
\end{itemize}
The problem (and the subject of the mentioned improvements) of this approach resides in the fact that after the first weight update, the weights on the PSs and the workers will be different (except for the worker who communicated with the PSs last). This in turn creates an inconsistency between the weights used to compute the gradient (on the worker's side) and the weights which will be updated with such gradient on the PSs. This problem is often reffered to as \textit{gradient staleness}. Clearly, the larger the difference between the weights, the less accurate the update will be. If we assume that all $N$ workers have approximately the same processing speed, we can deduce that after $N$ iterations the PSs will receive gradients which are -on average- out of date by $N$ steps. This clearly has a large negative impact on convergence, when $N$ is large. We will focus on one particular attempt which has been made to limit this effect and is derived in the DC-ASGD algorithm. The method computes an approximated first-order correction to modify the gradients received by the PSs. But even though this approach mitigates the problem, it can only work when the distance between PSs' and worker's weights is relatively small.

In recent years, large-scale training was obtained by using different flavors of the most classic synchronous scheme, that is Synchronous SGD, in conjunction with decentralized communication. Again, even though many variants exist, the core mechanism is easy to summarize as follows:
\begin{itemize}
    \item at the beginning of the computation, every worker receives an exact copy of the weights
    \item when a worker has finished processing its mini-batch, it participates in a blocking all-reduce operation, where it shares the gradient it computed with all other workers
    \item at the end of the all-reduce, all workers possess the sum of the computed gradients, and they can use it to compute the same weight update
    \item every worker proceeds to process another batch
\end{itemize}
This scheme has been thoroughly explored, and has one only drawback, which resides in the blocking nature of the all-reduce operation: all workers have to wait for the slowest one (sometimes referred to as \textit{straggler}) before initiating the communication, and then they have to wait for the end of the communication to compute the update.

Decentralized communication can also be used for a particular form of asynchronous methods, which are known as \textit{stale-synchronous}. In stale-synchronous methods, workers are allowed to go out of sync by a maximum number of iterations (processed mini-batches), before waiting for other ones to initiate communication. The maximum number of iterations is called maximum staleness.

As we will see in the next section, our method is a stale-synchronous centralized of DC-ASGD, and in this work, we will only focus on the version with a maximum staleness of one.

\section{Algorithm}
\label{sec:dcdssgd}
Our algorithm is similar to the DC-ASGD method proposed in \cite{DBLP:journals/corr/ZhengMWCYML16}, with three main differences
\begin{itemize}
    \item it eliminates the need of a Parameter Server in favor of a decentralized communication scheme;
    \item it is stale-synchronous, and not fully asynchronous;
    \item weights computed by different workers are averaged.
\end{itemize}
In the following sections, we will explain why these differences result in a novel and improved approach, compared to existing algorithms.

\subsection{Problem Setting}
\label{sec:problem_setting}
We quickly review the problem of data-parallel training of a DNN. For this work, we will focus on DNNs trained as multi-dimensional classifiers, where the input is a sample, denoted by $\mathbf{x}$. The goal of training is to find a set of network weights $\mathbf{w}$ which minimizes a loss function
\begin{equation}
   L(\mathbf{w}) = \frac {1}{\left| \mathcal{X} \right|} \sum_{\mathbf{x} \in \mathcal{X}} l(\mathbf{w}, \mathbf{x})
\end{equation}
for a set of samples $\mathcal{X}$, where $l(\mathbf{x}, \mathbf{w})$  is the per-sample classification loss function (cross-entropy loss in our case). Instead of reporting the final value of the loss function, it is usual to derive a figure of merit, which has the benefits of being more understandable by humans and applicable to different loss functions. In our case, we will use the top-1 error rate, which is simply the rate of misclassified samples to the number of elements of $\mathbf{x}$. We will measure both the error obtained on the training data set and  on the validation data set.

We will employ a common version of the classic \textit{Mini-batch Stochastic Gradient Descent}, which is usually referred to as \textit{Stochastic Gradient Descent} (SGD), and solves the above mentioned minimization problem in an iterative way, following
\begin{equation}
    \mathbf{w}^{t+1} = \mathbf{w}^t - \eta \frac{1}{\left| \mathcal{B} \right|} \sum_{x \in \mathcal{B}} \nabla l({\mathbf{x}, \mathbf{w}^t})
\end{equation}
where $\mathcal{B}$ is a mini-batch, i.e. a subset of the training data set, and  $\left| \mathcal{B} \right|$ is the mini-batch size, which has been proven to be an important factor, determining how easily a network can be trained. We will adopt a simple version of the SGD algorithm, namely the so-called momentum SGD, in which a momentum term \cite{Qian99onthe} ensures that updates are damped, and allows for faster learning \cite{Qian99onthe}.

In the synchronous parallel version, SGD works exactly in the same way, with the only difference that each worker computes gradients locally on the mini-batch it processes, and then shares them with other workers by means of an all-reduce call.

\subsection{DC-ASGD}
\label{sec:dcasgd}
Since our algorithm is a variation of DC-ASGD, we will briefly outline its most important feature, that is, the delay compensation. As illustrated in Section~\ref{sec:advantages}, gradient staleness reduces the convergence rate, because of the difference between the weights held by the worker and those held by the PSs. In DC-ASGD, the gradients are modified to take this difference into account. Basically, the idea is to apply a first-order correction to the gradients, so that they are approximately equal to those which would have been computed using the PSs' copy of the weights. If the Hessian matrix computed at $\mathbf{w}_i$, here denoted by $\mathbf{H}_i$, was known, one could compute the corrected gradients as
\begin{equation}
\label{eq:realhessiangrad}
    \mathbf{g}_{PS} = \mathbf{g}_i + \mathbf{H}_i \cdot (\mathbf{w}_{PS}-\mathbf{w}_i) + \mathcal{O}((\mathbf{w}_{PS}-\mathbf{w}_{i})^2)\cdot \mathbf{I}_n
\end{equation}
where $\mathbf{w}_i$ are the weights used by the $i^{\text{th}}$ worker,  $\mathbf{w}_{PS}$ are those held by the PS, and $\mathbf{I}^n$ is a vector with all $n$ components equal to one, with $n$ being the dimension of the weights. The quadratic error term $\mathcal{O}((\mathbf{w}_{PS}-\mathbf{w}_{i})^2)\cdot \mathbf{I}_n$ comes directly from the Taylor expansion used to derive this result, and we will denote it as $\mathcal{R}$ for the rest of this work. In principle, the Hessian matrix could be computed analytically, but the product of its approximation (known as pseudo-Hessian) $\widetilde{\mathbf{H}}$ with a vector $\mathbf{v}$ is computationally convenient to compute as 
\begin{equation}
    \widetilde{\mathbf{H}}_i \mathbf{v} = \mathbf{g}_i \odot \mathbf{g}_i \odot \mathbf{v}
\end{equation}
where $\odot$ represents the Hadamard (or component-wise) product.
Thus, we can rewrite \ref{eq:realhessiangrad} as
\begin{equation}
\label{eq:pseudohessiangrad}
    \mathbf{g}_{PS} \approx
    \mathbf{g}_i + \mathbf{g}_i \odot \mathbf{g}_i \odot (\mathbf{w}_{PS}-\mathbf{w}_i) + \mathcal{R}.
\end{equation}
Removing the error term and adding a variance control parameter $\lambda\in\mathbb{R}$ as defined in \cite{DBLP:journals/corr/ZhengMWCYML16}, we obtain the final form of the equation as
\begin{equation}
\label{eq:psgrad}
    \mathbf{g}_{PS} \approx \mathbf{g}_i + \lambda  \mathbf{g}_i \odot \mathbf{g}_i \odot (\mathbf{w}_{PS}-\mathbf{w}_i)
\end{equation}
which is the one we base our algorithm on.

\subsection{DC-S3GD}
In our centralized setting, there is no PS, but since we implement a stale-synchronous method, workers can be expected to be out of sync. In fact, the main idea of our approach is to allow for communication and computation to run in parallel, thus diminishing communication's impact on the total training run time. To allow for this, we make use of the non-blocking all-reduce function which is part of the MPI standard, i.e. \texttt{MPI\_Iallreduce}.

We now describe our method, which is also illustrated in Algorithm~\ref{algo:dcs3gd}. We stress the fact that all processing units will act as identical workers, only fed with different data. 
The only hyper-parameters we will need to set are the learning rate $\eta$, the momentum $\mu$, and the variance control parameter~$\lambda$.

At the beginning of the computation, each worker receives the same set of initial weights $\bar{\mathbf{w}}^0$ and a different mini-batch, which it processes to obtain a set of gradients $\mathbf{g}_i$, where the bar over $\mathbf{w}$ stresses the fact that the same value is held by all workers, the subscript $i$ denotes the worker index, and the superscript $0$ denotes the iteration. We will drop the superscripts when possible, to keep the notation concise.

Based on $\mathbf{g}_i$, the worker uses a function $\mathbf{U}(\mathbf{g}_i, \eta, \mu)$ to compute the update to its local weights. We denote the update as $\Delta \mathbf{w}_i^t$ and all workers will share their local update with the others, by starting a non-blocking all-reduce operation.

While the all-reduce operation is progressing, the worker updates its local copy of the weights:
\begin{equation}
    \mathbf{w}_i^{t+1} = \bar{\mathbf{w}}^{t} + \Delta \mathbf{w}_i^t
\end{equation}
and proceeds to process the next mini-batch, in order to compute new gradients $\mathbf{g}_i$.
After having processed the mini-batch, all workers wait for the all-reduce operation to complete. In our implementation, the completion is checked by means of a call to \texttt{MPI\_Wait}. After completion, each worker possesses an identical copy of $\overline{\Delta}\mathbf{w}$, that is the sum of all workers' updates of the previous iteration.

At this point, we can compute the average of the weights held by each worker, as
\begin{equation}
    \bar{\mathbf{w}}^{t+1} = \frac{1}{N} \sum_i \bar{\mathbf{w}}^t + \Delta \mathbf{w}_i^t = 
    \bar{\mathbf{w}}^t + \frac{1}{N} \overline{\Delta} \mathbf{w}^t.
\end{equation}
Notice that in principle, there is no guarantee that the mean value of the weights is actually meaningful, but studies such as \cite{DBLP:journals/corr/abs-1803-05407} suggest that averaging different weights can lead to better minima.
The Euclidean distance from the weights possessed by the $i^\text{th}$ worker to the average weights is
\begin{equation}
\begin{split}
    \mathbf{D}_i &= \bar{\mathbf{w}}^{t+1} - \mathbf{w}_i^{t+1} \\
    &= \bar{\mathbf{w}}^t + \frac{1}{N} \overline{\Delta} \mathbf{w}_i^t - \left(\bar{\mathbf{w}}^{t} + \Delta \mathbf{w}_i^t\right) \\
    &= \frac{1}{N} \overline{\Delta} \mathbf{w}_i^t  - \Delta \mathbf{w}_i^t
\end{split}
\end{equation}
Knowing this distance, each worker could replace its own copy of the weights with the average ones, but this is actually not needed. More importantly, by using a modified version of \ref{eq:psgrad}, the local gradient can be corrected and used to compute a local update that can be applied to the average weights. The correction equation becomes
\begin{equation}
      \widetilde{\mathbf{g}}_i = \mathbf{g}_i + \lambda_i \mathbf{g}_i ⊙ \mathbf{g}_i ⊙ \mathbf{D}_i
\end{equation}
and thus the new update can be computed as 
\begin{equation}
    \Delta \mathbf{w}_i = \mathbf{U}(\widetilde{\mathbf{g}}_i, \eta, \mu).
\end{equation}
and immediately shared with the other workers, by means of a new non-blocking all-reduce call.
Each worker will update its weights following
\begin{equation}
    \mathbf{w}_i = \mathbf{w}_i + \mathbf{D}_i + \Delta \mathbf{w}_i
\end{equation}
where we first move weights to the average value and update them as a single operation. At this point, each worker can start a new iteration, by proceeding to process the next mini-batch. 

A description of how $\lambda_i$ is computed at each iteration is given in \ref{sec:hyperparametersettings}.

\vspace{4 mm}
\begin{algorithm}[t]
\SetKwInput{KWInit}{Initialize}
\caption{DC-S3GD for $N$ workers}
\label{algo:dcs3gd}
\KwIn{step $\eta$, momentum $\mu$, variance control parameter $\lambda_0$}
\KWInit{weights $\mathbf{w}_i = \bar{\mathbf{w}}$}
$\mathbf{g}_i = \nabla l (\mathbf{w}_i)$\;
$\Delta \mathbf{w}_i = \mathbf{U}(\mathbf{g}_i, \eta, \mu)$ \;
$\mathbf{w}_i = \mathbf{w}_i + \Delta \mathbf{w}_i $ \;
 \For{t < max\_iterations}{
  \texttt{MPI\_Iallreduce}($\Delta \mathbf{w}_i$)\tcp*[l]{non-blocking} 
  $\mathbf{g}_i = \nabla l (\mathbf{w}_i)$\;
  $\overline{\Delta} \mathbf{w}$ = \texttt{MPI\_Wait}()\tcp*[l]{blocking}
  $\mathbf{D}_i = \frac{1}{N}\overline{\Delta} \mathbf{w} - \Delta \mathbf{w}_i$\;
  $\widetilde{\mathbf{g}}_i = \mathbf{g}_i + \lambda_i \, \mathbf{g}_i ⊙ \mathbf{g}_i ⊙ \mathbf{D}_i$\;
  $\Delta \mathbf{w}_i = \mathbf{U}(\widetilde{\mathbf{g}}_i, \eta, \mu)$\;
  $\mathbf{w}_i = \mathbf{w}_i + \mathbf{D}_i + \Delta \mathbf{w}_i $\;
 }
\end{algorithm}

\subsection{Advantages and Disadvantages of the proposed Method}
We compare the proposed approach to two methods described in \ref{sec:advantages}, SSGD and DC-ASGD.
\subsubsection{Comparison to SSGD}
The main advantage over SSGD resides in the fact that communication costs are (at least partially) hidden in our approach. We can approximate the time taken by SSGD to complete an iteration over a mini-batch $\mathcal{B}$ over $N$ nodes as
\begin{equation}
    t_{SSGD} = t_C(\mathcal{B}) + t_{ARed}(\mathbf{g}, N)
\end{equation}
where $t_C(\mathcal{B})$ is the time it takes a worker to process the mini-batch (including feed-forward and back-propagation phases), and $t_{ARed}(\mathbf{g}, N)$ is the time taken by the all-reduce call to reduce the gradients $\mathbf{g}$ across all nodes. For our method, a similar approximation can be made, and it yields
\begin{equation}
    t_{DC-S3GD} = \max(t_C(\mathcal{B}), t_{ARed}(\mathbf{g}, N))
\end{equation}
which is an obvious consequence of the fact that the computation and all-reduce operations run concurrently in our setting.
\subsubsection{Comparison to DC-ASGD}
Similarly to the results derived in the previous section, we can define an approximation to the run-time of a DC-ASGD iteration, denoted by $t_{SSGD}$, as
\begin{equation}
    t_{DC-ASGD} = t_C(\mathcal{B}) + t_{W2PS}(\mathbf{g}, N)
\end{equation}
where $t_{P2P}(\mathbf{g}, N)$ is the total time needed by a worker to push its gradients to the PS and obtain the updated weights. Clearly, this time also includes time spent by the worker, waiting for the PS to receive the gradients. Therefore, even though it is true that in DC-ASGD fast workers do not have to wait for stragglers, it is also true that run-time depends heavily on the network and on the capability of PSs.
As mentioned in \ref{sec:advantages}, DC-ASGD's convergence decreases for increasing numbers of workers. This is because the Euclidean distance between the workers' and the PSs' weights, $\mathbf{w}_{PS}-\mathbf{w}_i$, is proportional to $N$. In our method, the distance used to compute the correction is that between workers' and average weights, which we expect to grow more slowly w.r.t. $N$.

\section{Experiments}
\begin{table*}[htbp]
\begin{center}
\begin{tabular}{|l|c|c|c|c|c||c|}
\hline
\textbf{Network}	&$\left|\mathcal{B}\right|$	&\textbf{\#Nodes} & 
\textbf{Train Accuracy} & \textbf{Val. Accuracy} & \textbf{Speed [img/sec]} & Reference Val. Acc.\\
\hline
ResNet-50	&16k	&32	& 80.7\%	&77.5\%  & 2078 & 75.3\% \cite{you2017imagenet}, SSGD \\ 
\hline
ResNet-50	&32k	&32	&	80.3\% &77.4\%  & 2144 & 75.4\% \cite{you2017imagenet}, SSGD \\
\hline
ResNet-50	&32k	&64	& 78.5\%	&77.2\% & 3815 & 75.4\% \cite{you2017imagenet}, SSGD \\
\hline
ResNet-50	&64k	&64	& 76.6\%	&75.6\%  & 4245 & 76.2\% \cite{DBLP:journals/corr/abs-1807-11205}, SSGD\\
\hline
ResNet-50	&64k	&128 & 75.6\%	&75.1\% & 7340 & 76.2\% \cite{DBLP:journals/corr/abs-1807-11205}, SSGD \\
\hline
ResNet-50	&128k	&128 & 70.0\%	&69.7\%  & 8201 & 75.0\% \cite{osawa2018largescale}, K-FAC \\
\hline
ResNet-101	&64k &64	& 78.3\%		&77.2\% & 2578 &\\
\hline
ResNet-152	&32k	&64 & 80.9\%	&78.7\% & 1768 &\\
\hline
VGG-16	&16k	&64 & 63.03\%	&69.2\% &1206 &\\
\hline
\end{tabular}
\end{center}
\caption{Average validation accuracy and processing speed for parallel training of CNNs with DC-S3GD. The last columns shows reference results for the training of ResNet-50 with synchronous methods for the same batch size.}
\label{tab:results}
\end{table*}
\label{sec:results}

In this section, we first describe how we set training hyper-parameters, and then we report results obtained by training four standard CNNs on the ImageNet-1k data set.
\subsection{Hyper-parameter Settings and Update Schedules}
\label{sec:hyperparametersettings}
As mentioned in \ref{sec:problem_setting}, to train CNNs, we employed a data-parallel version of SGD with momentum. For each network, we set the momentum $\mu$ to the value used to obtain the state-of-the-art results, and we keep it constant for the whole training, which consisted in 90 full epochs.
For the learning rate $\eta$, we first define the theoretical learning rate as 
\begin{equation}
    \eta_{theo} = N \eta_{sn}
\end{equation}
where $N$ is the number of workers, as usual, and $\eta_{sn}$ is the learning rate for single-node training: for ResNet cases, we used as reference a learning rate of 0.1 for a batch-size of 256 samples. This is standard practice, and it seems to give stable results for our setting. For VGG, the base learning rate was 0.02.
Another standard approach is to define a learning rate schedule. In our case, we adopted an iteration-dependent (and not \textit{epoch-dependent}) schedule with linear warm-up and linear decrease. The length of the warm-up phase was initially defined as half of the total iterations, but we found empirically that after 15 epochs, the training error would reach a plateau (for all batch sizes up to 64k samples), and thus we stopped the warm-up phase at the reached learning rate, and we initiated a longer linear decrease phase, which would run until the end of the training. For the case of 128k samples, the plateau was reached after 20 epochs. Identification of the plateau was done by direct observation, but we believe it could easily be automated, by e.g. checking for training error reduction every five epochs during the warm-up phase.

To reduce over-fitting, weight decay was applied to all weights, with the exception of those belonging to batch normalization layers. This technique has given the best results, and the reasoning behind it can be found in \cite{DBLP:journals/corr/abs-1807-11205}. 
Since this kind of normalization reduces weights by a constant fraction, when the learning rate is very little (as it can happen in our case, when $t$ is very close to 0 or to $max\_iterations$), the weight decay can become larger than the update, therefore blocking convergence. To mitigate this problem, we decided to apply the same schedule we used for learning rate, also for the weight decay parameter. To compensate for the smaller effective regularization, we also multiply the weight decay hyper-parameter by a constant factor $k$. We find that $k=2.3$ gives us the best results. This factor was applied to the weight decay hyper-parameter value usually adopted in the literature, namely 0.0001 for ResNet topologies and VGG-16.

By stopping the warm-up phase early, we reach only a small fraction of the maximum step length (e.g. one third for a 15-epoch warm-up), and we note that the pseudo-Hessian correction term is very small compared to the computed gradients. We investigated possible correction re-scaling techniques, and we found that the best result was to add 0.5\% to the validation accuracy, when the step reached the end of the warm-up phase. We think that this correction term would have a larger influence for larger learning rates. The parameter $\lambda_i$, which is used to control the variance introduced by correction step \cite{DBLP:journals/corr/ZhengMWCYML16}, was empirically found to give the best results when dynamically set as
\begin{equation}
    \lambda_i = \frac{\lambda_0 \left\lVert \mathbf{g}_i \right\rVert }{\left\lVert \mathbf{g}_i  ⊙ \mathbf{g}_i ⊙ \mathbf{D}_i\right\rVert }
\end{equation}
with $\lambda_0=0.2$.

\subsection{Hardware and Software Configuration}
We ran our experiments on a Cray XC system.
Every node was equipped with two 24-core Intel Skylake processors with a clock speed of 2.4 GHz and nodes were connected through Cray Aries with dragonfly topology. The use of CPUs only, which is in contrast with the more standard usage of a GPU-cluster, allowed us to explore very large local mini-batch sizes (up to 1024 samples per local mini-batch).
As a toolkit, we used a modified version of MXNet \cite{MXNet}, in conjunction with the Intel MKL-DNN libraries \cite{mkl-dnn}. We chose to use MXNet because it offered an easy way to implement our algorithm: we modified the original Key-Value Store (KV Store), which is used to update weights after each iteration, so that it included the needed mechanics and MPI code.
The MPI implementation was \texttt{Cray-mpich}. 
The source code can be made available upon direct request to the author.

\subsection{Results}
We report results obtained by training ResNet-50, ResNet-101, ResNet-152, and VGG-16 on the ImageNet-1k data set.

\begin{figure*}[htbp]
    \centering
    \includegraphics[width=\textwidth]{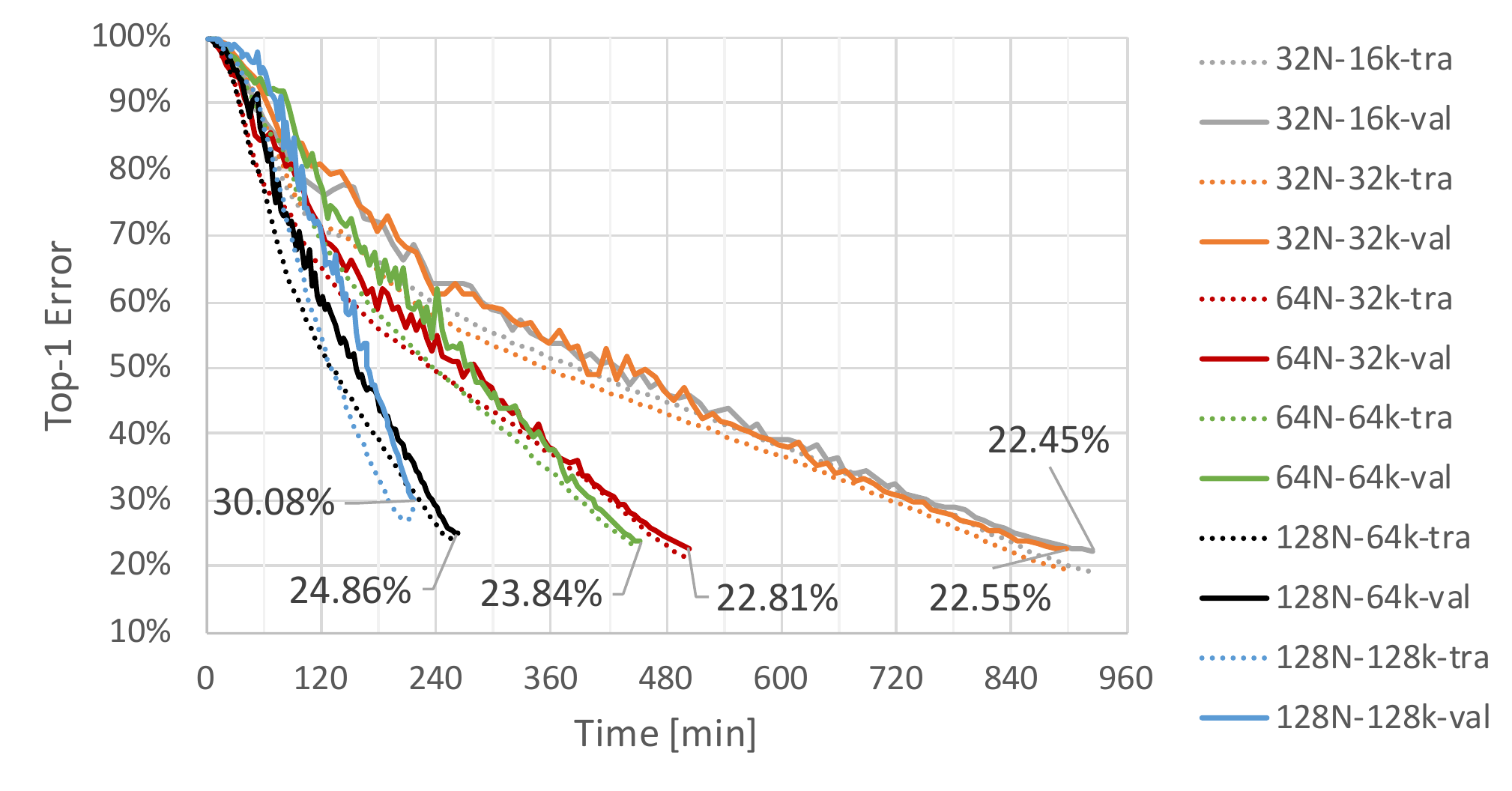}
    \caption{ResNet-50 Top-1 training and validation errors for different combinations of node count $N$ and aggregate mini-batch size.}
    \label{fig:resnet50}
\end{figure*}

\subsubsection{ResNet-50}
As training ResNet-50 has become a reference benchmark, we investigated performances of our method on such problem, for different settings. To maximize CPU usage, and to exploit the large memory available on CPU nodes, we use a local mini-batch size of 512 or 1024 samples. From the achieved accuracy values, shown in Table~\ref{tab:results}, it can be seen that we manage to reach state-of-the-art accuracy on up to 64 nodes, with a batch size of 32k samples: the total training time, not considering network setup, is of 503 minutes. Keeping the number of nodes at 64 and using a larger batch size results in a slight loss of accuracy and a speed-up of ~10\%. Running the parallel training on 128 nodes, we still reach a reasonable accuracy for a total mini-batch size of 64k samples, in ~260 minutes: in comparison to \cite{MLPerf}, where the target accuracy was 74.9\%, we clearly outperform the best results obtained on CPUs, even accounting for the difference between total execution and training time, which never exceeded 10 minutes. From the reported results, we can see that employing a larger batch size on 128 nodes results in a large loss of accuracy. In Figure~\ref{fig:resnet50}, top-1 error for full training of ResNet-50 networks is shown. For each combination of node count and aggregate batch size, we plot the results of the training run which reached the lowest validation error.

\subsubsection{Other Architectures}
Table~\ref{tab:results} lists the results we obtained training other CNNs. It is clear that we are able to reach state-of-the-art accuracy for all ResNet topologies. More importantly, our method is also capable of training VGG-16 with a mini-batch size of 16k samples, even though this is known to be a difficult task \cite{DBLP:journals/corr/abs-1708-03888}.

In order to fairly assess our method's performances, we did not adapt the hyper-parameters for the different topologies. The only tuning we performed, was to extend the warm-up phase to 20 epochs (thus, two ninth of the total training) when running on 64 or 128 nodes.

\section{Conclusions}
\label{sec:future_work}
In this work, we proposed a new algorithm for distributed training, named DC-S3GD, which allows for the overlap of computation and communication by averaging in the parameter space (weights) and applying a first-order correction to gradients. We showed that this approach can achieve state-of-the-art results for parallel DL training. 

Many aspects could be improved, for example, more sophisticated methods, like LARS \cite{DBLP:journals/corr/abs-1708-03888}, or Adam \cite{Adam}, could be used as local optimizers.

Another possible enhancement would be to allow more out-of-sync minimization steps to be taken by local optimizers, and to see how this influences performances, in terms of time-to-accuracy.

To reduce the error introduced in the correction step, the pseudo-Hessian could be replaced by an analytical version of the Hessian matrix.

In terms of maximum achieved accuracy, we ran some preliminary tests with a larger number of iterations, and in some cases, extending the training to 100 or 120 epochs could improve the accuracy of 0.2-0.8\%, even for the case of 128k samples per batch.

We believe this approach could also be applied to train neural networks of other types, such as those used for Natural Language Processing, or Reinforcement Learning, if a data-parallel scheme can be adopted.

\section{References}
\vspace{1\baselineskip}
\bibliographystyle{abbrv}

\begingroup
\renewcommand{\section}[2]{}%
\begin{footnotesize}
\bibliography{bibliography.bib,alrigazzi_bibsonomy.bib,sparse_bib.bib,DeepLearning.bib}

\begin{thebibliography}{10}

\bibitem{tensorflow2015-whitepaper}
M.~Abadi, A.~Agarwal, P.~Barham, E.~Brevdo, Z.~Chen, C.~Citro, G.~S. Corrado,
  A.~Davis, J.~Dean, M.~Devin, S.~Ghemawat, I.~Goodfellow, A.~Harp, G.~Irving,
  M.~Isard, Y.~Jia, R.~Jozefowicz, L.~Kaiser, M.~Kudlur, J.~Levenberg,
  D.~Man\'{e}, R.~Monga, S.~Moore, D.~Murray, C.~Olah, M.~Schuster, J.~Shlens,
  B.~Steiner, I.~Sutskever, K.~Talwar, P.~Tucker, V.~Vanhoucke, V.~Vasudevan,
  F.~Vi\'{e}gas, O.~Vinyals, P.~Warden, M.~Wattenberg, M.~Wicke, Y.~Yu, and
  X.~Zheng.
\newblock {TensorFlow}: Large-scale machine learning on heterogeneous systems,
  2015.
\newblock Software available from tensorflow.org.

\bibitem{DBLP:journals/corr/abs-1802-09941}
T.~Ben{-}Nun and T.~Hoefler.
\newblock Demystifying parallel and distributed deep learning: An in-depth
  concurrency analysis.
\newblock {\em CoRR}, abs/1802.09941, 2018.

\bibitem{chen2015mxnet}
T.~Chen, M.~Li, Y.~Li, M.~Lin, N.~Wang, M.~Wang, T.~Xiao, B.~Xu, C.~Zhang, and
  Z.~Zhang.
\newblock Mxnet: A flexible and efficient machine learning library for
  heterogeneous distributed systems, 2015.
\newblock cite arxiv:1512.01274Comment: In Neural Information Processing
  Systems, Workshop on Machine Learning Systems, 2016.

\bibitem{MXNet}
T.~Chen, M.~Li, Y.~Li, M.~Lin, N.~Wang, M.~Wang, T.~Xiao, B.~Xu, C.~Zhang, and
  Z.~Zhang.
\newblock {MXN}et: {A} flexible and efficient machine learning library for
  heterogeneous distributed systems.
\newblock {\em CoRR}, abs/1512.01274, 2015.

\bibitem{mkl-dnn}
I.~Corporation.
\newblock \url{https://01.org/mkl-dnn}.

\bibitem{goyal2017accurate}
P.~Goyal, P.~Dollár, R.~Girshick, P.~Noordhuis, L.~Wesolowski, A.~Kyrola,
  A.~Tulloch, Y.~Jia, and K.~He.
\newblock Accurate, large minibatch sgd: Training imagenet in 1 hour, 2017.
\newblock cite arxiv:1706.02677Comment: Tech report (v2: correct typos).

\bibitem{Gupta7837841}
S.~{Gupta}, W.~{Zhang}, and F.~{Wang}.
\newblock Model accuracy and runtime tradeoff in distributed deep learning: A
  systematic study.
\newblock In {\em 2016 IEEE 16th International Conference on Data Mining
  (ICDM)}, pages 171--180, Dec 2016.

\bibitem{DBLP:journals/corr/abs-1803-05407}
P.~Izmailov, D.~Podoprikhin, T.~Garipov, D.~P. Vetrov, and A.~G. Wilson.
\newblock Averaging weights leads to wider optima and better generalization.
\newblock {\em CoRR}, abs/1803.05407, 2018.

\bibitem{DBLP:journals/corr/abs-1807-11205}
X.~Jia, S.~Song, W.~He, Y.~Wang, H.~Rong, F.~Zhou, L.~Xie, Z.~Guo, Y.~Yang,
  L.~Yu, T.~Chen, G.~Hu, S.~Shi, and X.~Chu.
\newblock Highly scalable deep learning training system with mixed-precision:
  Training imagenet in four minutes.
\newblock {\em CoRR}, abs/1807.11205, 2018.

\bibitem{keuper2015asynchronous}
J.~Keuper and F.-J. Pfreundt.
\newblock Asynchronous parallel stochastic gradient descent - a numeric core
  for scalable distributed machine learning algorithms, 2015.
\newblock cite arxiv:1505.04956.

\bibitem{Adam}
D.~P. Kingma and J.~Ba.
\newblock Adam: A method for stochastic optimization, 2014.
\newblock cite arxiv:1412.6980Comment: Published as a conference paper at the
  3rd International Conference for Learning Representations, San Diego, 2015.

\bibitem{krizhevsky2014weird}
A.~Krizhevsky.
\newblock One weird trick for parallelizing convolutional neural networks,
  2014.
\newblock cite arxiv:1404.5997.

\bibitem{Li:2018:VLL:3327345.3327535}
H.~Li, Z.~Xu, G.~Taylor, C.~Studer, and T.~Goldstein.
\newblock Visualizing the loss landscape of neural nets.
\newblock In {\em Proceedings of the 32Nd International Conference on Neural
  Information Processing Systems}, NIPS'18, pages 6391--6401, USA, 2018. Curran
  Associates Inc.

\bibitem{ma2017accelerated}
C.~Ma, M.~Jaggi, F.~E. Curtis, N.~Srebro, and M.~Takáč.
\newblock An accelerated communication-efficient primal-dual optimization
  framework for structured machine learning, 2017.
\newblock cite arxiv:1711.05305.

\bibitem{openai2018empirical}
S.~McCandlish, J.~Kaplan, D.~Amodei, and O.~D. Team.
\newblock An empirical model of large-batch training, 2018.
\newblock cite arxiv:1812.06162.

\bibitem{MLPerf}
MLPerf.
\newblock v0.5 training, closed division times.
\newblock \url{https://www.mlperf.org}.
\newblock Retrieved from www.mlperf.org, 1 June 2019. MLPerf name and logo are
  trademarks. See www.mlperf.org for more information.

\bibitem{osawa2018largescale}
K.~Osawa, Y.~Tsuji, Y.~Ueno, A.~Naruse, R.~Yokota, and S.~Matsuoka.
\newblock Large-scale distributed second-order optimization using
  kronecker-factored approximate curvature for deep convolutional neural
  networks, 2018.

\bibitem{Qian99onthe}
N.~Qian.
\newblock On the momentum term in gradient descent learning algorithms, 1999.

\bibitem{smith2017bayesian}
S.~L. Smith and Q.~V. Le.
\newblock A bayesian perspective on generalization and stochastic gradient
  descent, 2017.
\newblock cite arxiv:1710.06451Comment: 13 pages, 9 figures. Published as a
  conference paper at ICLR 2018.

\bibitem{DBLP:journals/corr/WangGCLY17}
F.~Wang, X.~Gao, G.~Chen, W.~Li, and J.~Ye.
\newblock {IS-ASGD:} importance sampling accelerated asynchronous {SGD} on
  multi-core systems.
\newblock {\em CoRR}, abs/1706.08210, 2017.

\bibitem{you2017large}
Y.~You, I.~Gitman, and B.~Ginsburg.
\newblock Large batch training of convolutional networks, 2017.
\newblock cite arxiv:1708.03888.

\bibitem{DBLP:journals/corr/abs-1708-03888}
Y.~You, I.~Gitman, and B.~Ginsburg.
\newblock Scaling {SGD} batch size to 32k for imagenet training.
\newblock {\em CoRR}, abs/1708.03888, 2017.

\bibitem{you2017imagenet}
Y.~You, Z.~Zhang, C.-J. Hsieh, J.~Demmel, and K.~Keutzer.
\newblock Imagenet training in minutes, 2017.

\bibitem{DBLP:journals/corr/ZhengMWCYML16}
S.~Zheng, Q.~Meng, T.~Wang, W.~Chen, N.~Yu, Z.~Ma, and T.~Liu.
\newblock Asynchronous stochastic gradient descent with delay compensation for
  distributed deep learning.
\newblock {\em CoRR}, abs/1609.08326, 2016.

\end{thebibliography}
\end{footnotesize}
\endgroup

\end{document}